\documentclass{article} 
\usepackage[submission]{colm2025_conference}

\usepackage{microtype}
\usepackage{hyperref}
\usepackage{url}
\usepackage{booktabs}
\usepackage{graphicx}
\usepackage{makecell}
\usepackage{csquotes}
\usepackage{caption}
\usepackage{wrapfig}
\usepackage{makecell}

\usepackage{lineno}

\newif\ifcolmfinal
\colmfinaltrue

\definecolor{darkblue}{rgb}{0, 0, 0.5}
\hypersetup{colorlinks=true, citecolor=darkblue, linkcolor=darkblue, urlcolor=darkblue}

\title{The Zero Body Problem: Probing LLM Use of Sensory \\Language}

\author{Rebecca M.\ M.\ Hicke \\
Department of Computer Science\\
Cornell University\\
Ithaca, NY, USA \\
\texttt{rmh327@cornell.edu} \\
\And
Sil Hamilton \& David Mimno \\
Department of Information Science \\
Cornell University \\
Ithaca, NY, USA \\
\texttt{\{srh255, mimno\}@cornell.edu} \\
}

\begin{document}

\maketitle

\begin{abstract}
Sensory language expresses embodied experiences ranging from taste and sound to excitement and stomachache. 
This language is of interest to scholars from a wide range of domains including robotics, narratology, linguistics, and cognitive science.
In this work, we explore whether language models, which are not embodied, can approximate human use of embodied language.
We extend an existing corpus of parallel human and model responses to short story prompts with an additional 18,000 stories generated by 18 popular models.
We find that all models generate stories that differ significantly from human usage of sensory language, but the direction of these differences varies considerably between model families.
Namely, Gemini models use significantly more sensory language than humans along most axes whereas most models from the remaining five families use significantly less.
Linear probes run on five models suggest that they are capable of identifying sensory language.
However, we find preliminary evidence suggesting that instruction tuning may discourage usage of sensory language.
Finally, to support further work, we release our expanded story dataset.
\end{abstract}

\section{Introduction}

Sensory language makes use of descriptive words and phrases to evoke embodied experiences.
It encompasses language that appeals to senses like taste and sound and representations of physical actions.
The use of sensory and embodied language is of interest to scholars from many domains including robotics \citep{MADDEN2010180, Alomari_Duckworth_Hogg_Cohn_2017, Taniguchi17062016}, linguistics \citep{winter2019sensory, mondada2021language, lievers2018sensory}, cognitive science \citep{muraki2023insights, dove2022words, davis2021building, caballero2023sharing}, and narratology \citep{piper2024what, piper-bagga-2024-using, caracciolo2024bodies, herman2009basic, fludernik1996towards}.
Because large language models (LLMs) do not have embodied experiences, we might expect their use of sensory language to be consistently different from that of humans.
On the other hand, language models do not have \textit{any} experiences, and only learn from examples of human-written text, which includes sensory language. Therefore, they may replicate human linguistic patterns. In this work, we explore these hypotheses and probe whether LLMs mimic human levels of sensory language usage.

To do this, we extend an existing corpus of parallel human and GPT-3.5 responses to short story prompts \citep{huang2024gpt}. 
We select 1,000 prompts from the original dataset, randomly sample a human and GPT-3.5 response to each prompt, and then generate responses from an additional 18 language models from six prominent model families: Gemini, GPT, Llama, OLMo, Phi, and Qwen.
We then use two pre-existing lexicons from cognitive science \citep{lynott2020lancaster, brysbaert2014concreteness} to measure the strength of sensory language usage in all 20,000 texts along twelve axes: auditory, gustatory, haptic, interoceptive, olfactory, visual, foot/leg, hand/arm, head, mouth, torso, and concreteness. 
We finally perform pairwise comparisons between the strength of sensory content in human-written texts and texts written by each model. 

All nineteen models generated stories differing from human usage of sensory language in at least ten of the twelve sensory axes.
However, the direction of these differences varied considerably between model families.
In particular, Gemini models used significantly \textit{more} sensory language than humans along most axes, while most Llama, OLMo, Phi, and Qwen models used significantly \textit{less} sensory language.
The differences for the GPT models varied considerably between models and sensory axes.

We propose two possible explanations for this behavior.
First, LLMs may fail to learn (and therefore replicate) sensory language during pre-training.
However, we find that linear probes run on intermediate representations of human-written texts from five LLMs indicate that models \textit{are} able to recognize sensory language usage, at least for several axes. 
Second, model usage of sensory language may be influenced by instruction tuning.
An analysis of the sensory content in encouraged and discouraged model responses from one popular reinforcement learning from human feedback (RLHF) dataset \citep{bai2022training} supports this hypothesis. 
Specifically, we find a strong correlation between underuse of sensory language by models and over-representation of sensory language in discouraged model responses from the RLHF dataset.

LLMs' embodied language use provides a fascinating case study that has impact from abstract philosophy to practical human-model interactions.
This work demonstrates that LLMs do not approximate human use of embodied language when prompted for creative writing. 
It further  provides evidence that LLMs are able to represent sensory content but that instruction tuning may discourage its usage.
We appear to have discovered not a window but a mirror; we asked why LLMs don't use sensory language, and found it seems to be because we asked them not to.

\section{Related Work}

Considerable academic interest has recently been paid to LLMs' sensory knowledge. 
Much of this research has focused on probing models' abilities to replicate human-like sensory judgments for features like color \citep{kawakita2024, paik-etal-2021-world, marjieh2024}, sound \citep{siedenburg2023language}, and the perceptual strength of words along several sensory axes (including six studied in this paper) \citep{lee2025exploringmultimodalperceptionlarge}.
Other studies examine  whether embodied concepts like space and time \citep{gurnee2024language}, object sounds \citep{ngo-kim-2024-language}, implicit visual features \citep{jones-trott-2024-multimodal}, and color \citep{abdou-etal-2021-language} are represented in models' internal states. 
More work has investigated whether models can identify stimuli from human descriptions of sensory experiences \citep{zhong2024exploring, zhong2024sniffaispicyspicy}, whether humans understand model descriptions of sensory features \citep{zhong2024sniffaispicyspicy}, and whether models can reason about generic visual concepts \citep{zhang-etal-2022-visual} or the visual-auditory properties of language \citep{lee-lim-2024-language}. 
Finally, some papers inspect whether incorporating increased multi-modal or perceptual training improves models' performance on related tasks \citep{kennington-2021-enriching, li-etal-2024-groundinggpt}.

While the studies described above largely find that LLMs demonstrate at least some knowledge of sensory experience, to our knowledge no work has examined whether LLMs' use of sensory \textit{language} is human-like or if their descriptions of sensory experiences in non-explicitly perceptual tasks mimics human language.

\section{Data}

To create a comparative corpus of human and model short stories, we build on the GPT-WritingPrompts dataset \citep{huang2024gpt}.
This dataset is itself an extension of the original WritingPrompts dataset \citep{fan-etal-2018-hierarchical} which contains 303,358 stories written by users from \texttt{r/WritingPrompts}\footnote{\url{https://www.reddit.com/r/WritingPrompts/}} in response to 97,222 creative writing prompts.
Because this dataset was collected before and released in 2018, we assume there is little to no computationally generated content in the human responses. 
\citet{huang2024gpt} expanded this dataset by generating 206,226 responses to 97,219 of those prompts from GPT-3.5-turbo.
They used a temperature of 0.95 and two system prompts: the `author' prompt (``You are an award winning creative short story writer.'') and the `reddit' prompt (``You’re writing a Reddit story and you want other reddit users to like and upvote your story.'').

We further extend this dataset to new model families by randomly selecting a subset of 1,000 prompts from the GPT-WritingPrompts dataset.
We reject all `prompts' that are not creative writing prompts (such as community announcements) and prompts that do not pass GPT-4o's safety filters.
We then sample a random human and GPT-3.5 response to each of these prompts, ensuring that each human response is a story and not a comment or other forum content.
We then generate additional responses to the 1,000 prompts from each of the following 18 models:

\begin{itemize}
    \item \textbf{Gemini}: 1.5 --- Flash-8b, Flash \citep{team2024gemini}, 2.0 --- Flash-Lite, Flash \citep{gemini2.0}
    \item \textbf{GPT}: 4o \citep{openai2024gpt4ocard}
    \item \textbf{Llama}: 3.1 --- 8b, 3.2 --- 1b, 3b, 11b (vision), 3.3 --- 70b \citep{grattafiori2024llama}
    \item \textbf{OLMo 2}: 7b, 13b \citep{olmo20242}
    \item \textbf{Phi 4}: base, mini \citep{abdin2024phi}
    \item \textbf{Qwen 2.5}: 0.5b, 1.5b, 3b, 7b \citep{qwenQwen25TechnicalReport2025}
\end{itemize}

We select the instruction-tuned variant of each model to enable convenient story generation and follow \citet{huang2024gpt} in setting temperature to 0.95.
For each model, we randomly apply the `author' system prompt when generating responses to 500 prompts and the `reddit` prompt for the remaining 500.
The Gemini and GPT models are accessed using the Google and Azure OpenAI APIs respectively. 
The remaining models were all accessed via HuggingFace.
We release the additional generations for each prompt \href{https://anonymous.4open.science/r/sensorylanguage-6AC9/}{here}. 

In order to measure sensory word usage, we further draw on two datasets from cognitive linguistics: a sensorimotor lexicon \citep{lynott2020lancaster} and a concreteness lexicon, where concreteness is defined as ``the degree to which the concept denoted by a word refers to a perceptible entity'' \citep{brysbaert2014concreteness}.
The sensorimotor lexicon includes information on eleven sensory axes, six of which are perceptual modalities (haptic, auditory, olfactory, gustatory, visual, and interoceptive) and five of which are action effectors, or body parts which respond to a stimulus (mouth/throat, hand/arm, foot/leg, head excluding mouth/throat, and torso).
The mean ratings along these axes were generated for 37,058 English-language lemmas by 3,500 native English speakers on Mechanical Turk who ranked lemmas on a scale of 0 to 5, where 0 represented lemmas ``not experienced at all with that sense/action'' for the sensorimotor axes or abstract terms and 5 represented lemmas ``experienced greatly with that sense/action'' or concrete terms \citep{lynott2020lancaster}.
Further information about these datasets and the data collection process can be found in the original articles.

\begin{table}[t]
    \small
    \centering
    \begin{tabular}{llc}
        \toprule
        \textbf{Sensory Axis} & \textbf{Highly Sensory Sample} & \textbf{Sensory Score} \\
        \midrule
        Auditory & The air buzzed with polyphonic chatter. & 0.84 \\
        \midrule
        Gustatory & He buttered a colossal slice of bread. & 0.87 \\
        \midrule
        Haptic & Twirling the pen between his fingers. & 0.75 \\
        \midrule
        Interoceptive & The capacity to love others richly begins. & 0.41 \\
        \midrule
        Olfactory & Her sweat bore the fragrance and sweetness of fruit. & 0.78 \\
        \midrule
        Visual & The sun peeked over the horizon. & 0.86 \\
        \midrule
        Foot -- Leg & She settled into her stride. & 0.69 \\
        \midrule
        Hand -- Arm & A muscle-bound wizard was doing bicep curls. & 1.14 \\
        \midrule
        Head & He didn’t need to blink. & 0.62 \\
        \midrule
        Mouth & A knowing smile played on her lips. & 0.54 \\
        \midrule
        Torso & My lungs filled the void left in his ribcage. & 0.65 \\
        \midrule
        Concreteness & Cold air snuck past my tattered clothing. & 1.23 \\
        \bottomrule
    \end{tabular}
    \caption{Examples of the texts that achieve high scores along the twelve axes of sensory language. Each text is paraphrased from an output given by GPT 3.5.}
    \label{tab:exampleTexts}
\end{table}

\section{Methods}

\paragraph{Measuring Sensory Language}
Our goal is to measure sensory language usage in the produced texts along each of the twelve axes outlined above.
We would like to represent the relative sensory content of each story, giving more weight to the sensory contributions of less common words and down-weighting the impact of frequently used (or stop) words.
To do this, we first tokenize and lemmatize each text in our dataset using \texttt{spaCy} \citep{honnibal2020spacy}. 
We then use \texttt{scikit-learn} to find the inverse document frequency (IDF) values for each lemma in our corpus \citep{scikit-learn}.
We calculate the IDF values for each lemma from the dataset consisting of human and model responses to each of the 1,000 selected prompts (20,000 documents total). 
Next, we normalize the IDF values to fall between zero and one by dividing each value by the largest IDF value in the dataset.

We then use the sensory lexicons in concert with the normalized IDF values to create normalized sensory scores along each sensory axis for each story.
To create these scores, we first look up every lemma in a text in the sensory lexicons. 
If the lemma is included in the lexicons, we multiply its sensory value along each axis by its normed IDF value and add the resulting numbers to the summed axis scores for the story. 
Finally, we normalize the summed sensory scores for a text by dividing by the number of lemmas used to calculate the score; that is, the number of lemmas in the text found in the sensory lexicons. 
This essentially mimics norming the scores by a story's word count, but avoids skewing the results if certain `authors' use terms not found in the lexicon more frequently.
This leaves us with twelve normalized strength scores for each story representing the magnitude of sensory language usage along each axis.
Examples of texts that rank highly along each axis with their corresponding sensory strength scores can be found in Table \ref{tab:exampleTexts}.

\paragraph{Comparing Sensory Language}
We can now use the numerical representations of each text's sensory language usage to compare responses by humans and each model. 
We produce two numerical comparisons of human and model sensory language usage. 
First, for each axis we subtract the mean sensory strength of all the model's responses from the mean sensory strength of all the human responses.
A negative difference in averages means that the model used more sensory language than the humans on average and vice versa.

Next, to determine whether the differences in sensory language usage are significant, we use a paired t-test.
Specifically, we compare the normalized sensory strengths of human and model responses to the same prompts along each axis.\footnote{Paired t-tests are implemented using \texttt{scipy}'s \texttt{ttest\_rel()} function.} 
We then report the resulting t-test statistics for each test, which provide information on the significance, strength, and direction of the differences in distribution means. 
Again, negative t-test scores indicate that the model used more sensory language than the human writers and vice versa.
All t-test statistics $\geq 1.96$ or $\leq -1.96$ are significant at $\alpha=0.05$. 

\paragraph{Distinguishing Model- and Human-Written Texts}
We would like to further probe whether model- and human-written texts can be distinguished using only the strength of their sensory language usage along all twelve axes of interest.
Having calculated the magnitude of sensory language usage for each text along each axis, we represent each story as a vector composed of each respective sensory strength. We then use these vectors to train and evaluate the ability of logistic regression models to distinguish between texts produced by humans and each LLM.

For every LLM, we train 100 logistic regression models to distinguish between stories written by humans and that LLM.
We reshuffle the train/test data splits for every model.
Each training dataset contains the responses to 500 randomly selected prompts (1,000 vectors) and the test dataset contains the responses to the remaining 500 prompts.
For each LLM, we report the average F1 score over all 100 logistic regression models and the standard deviation of all F1 scores.
Further, we find the importance of each sensory axis by permuting the input features\footnote{Implemented with \texttt{scikit-learn}'s \texttt{permutation\_importance} function.} and report the average importance of each feature.

\section{Comparing Sensory Language}
\label{sec:comparing sensory language}

\begin{figure*}[t]
    \centering
    \includegraphics[width=0.49\linewidth]{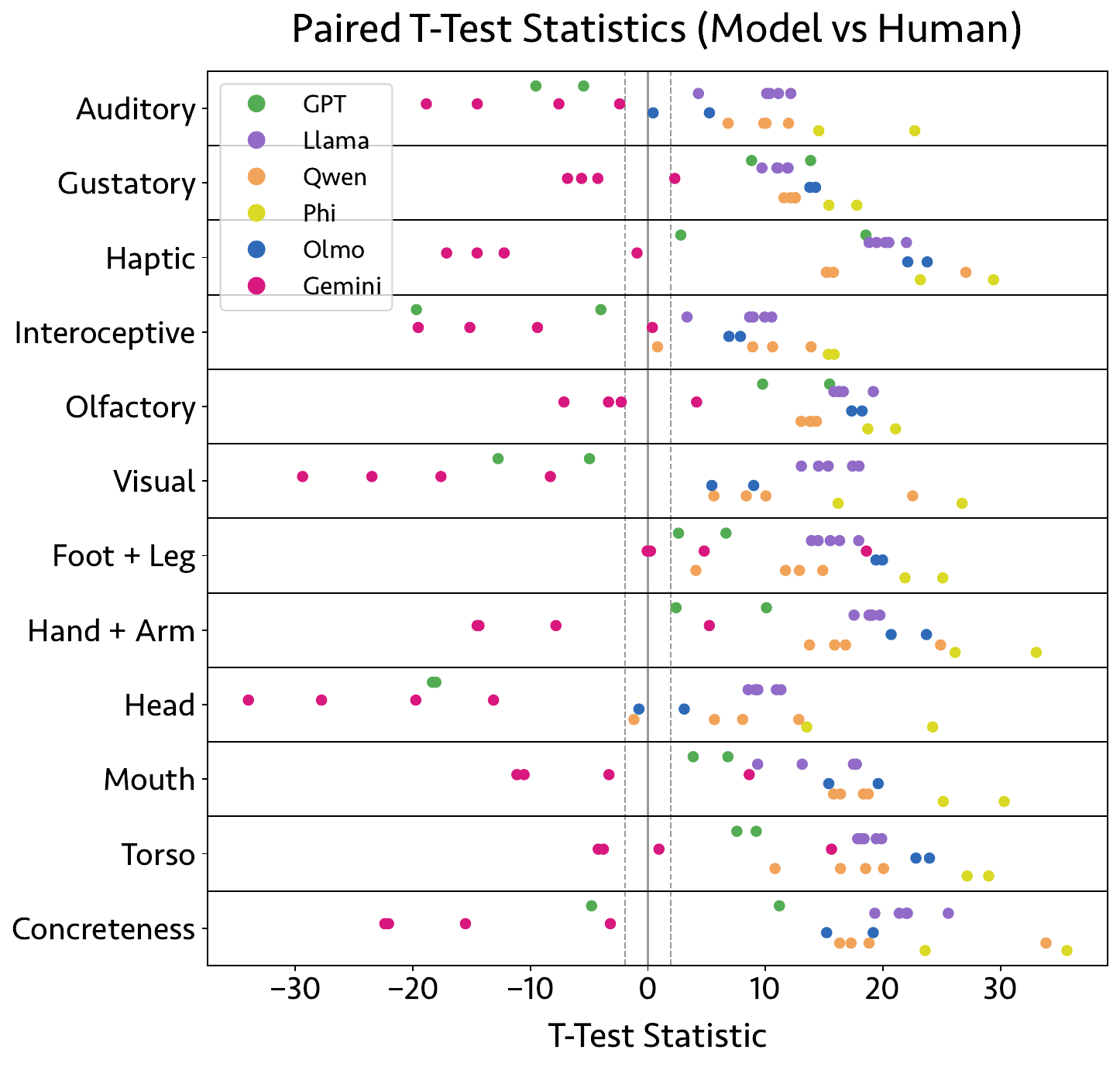}
    \hfill
    \includegraphics[width=0.49\linewidth]{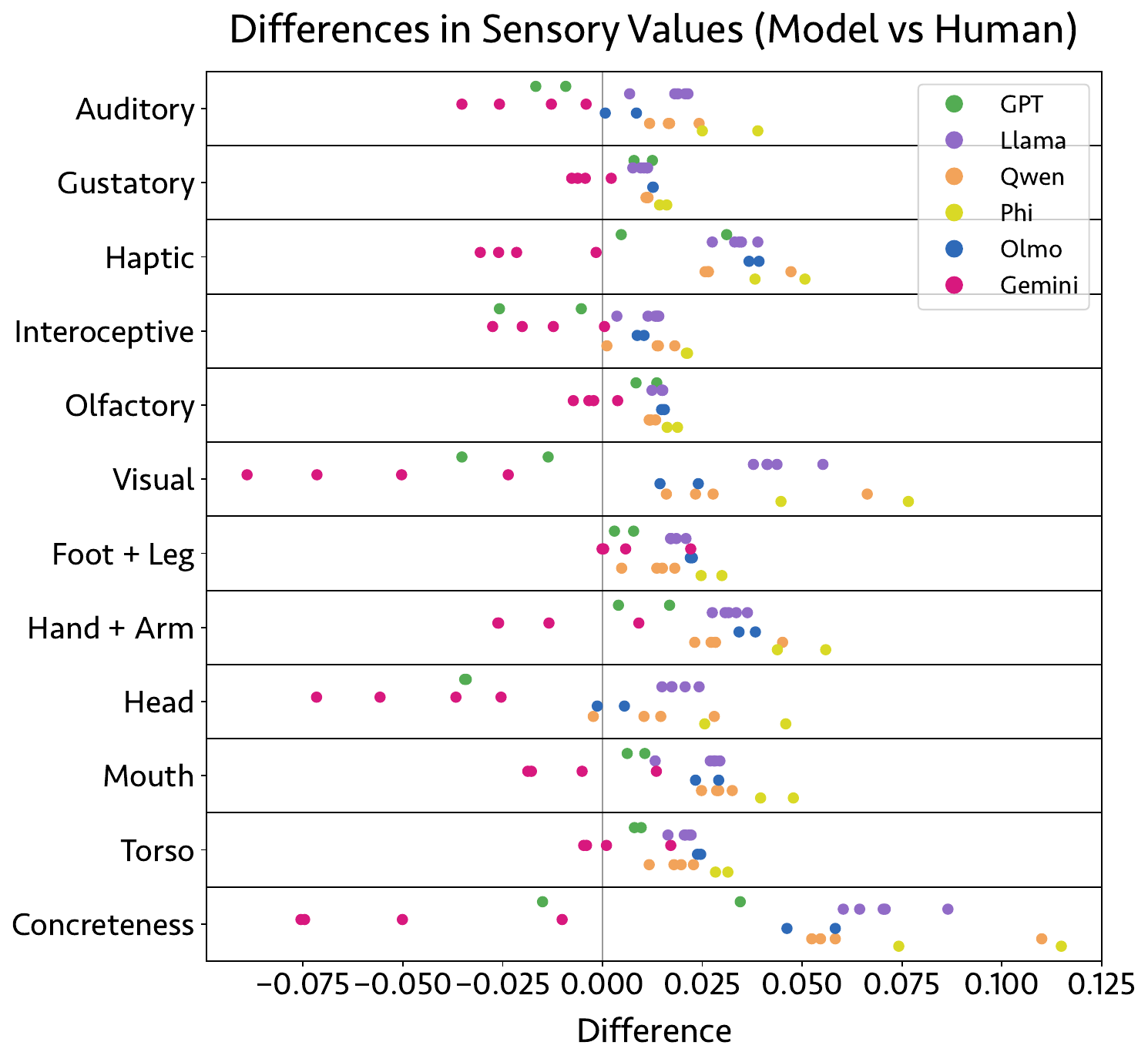}
    \caption{Comparisons of model and human use of sensory usage along each of the twelve axes for each model; values $<0$ indicate the model uses more sensory language and vice versa. The paired t-test statistics (left) and average differences (right) demonstrate significant differences in model and human language usage for most models along most axes. The dotted vertical lines on the left figure mark significance at $\alpha=0.05$; any t-test statistics $\geq1.96$ or $\leq -1.96$ represents a significant difference in model and human language usage.}
    \label{fig:sensoryAxes}
\end{figure*}

\begin{wraptable}{r}{0.4\textwidth}
    \small
  \centering
  \begin{tabular}{lc}
    \toprule
        \textbf{Axis} & \textbf{Avg. Strength} \\
        \midrule
        Auditory & 0.41 \\
        Gustatory & 0.10 \\
        Haptic & 0.30 \\
        Interoceptive & 0.29 \\
        Olfactory & 0.12 \\
        Visual & 0.68 \\
        Foot + Leg & 0.25 \\
        Hand + Arm & 0.38 \\
        Head & 0.57 \\
        Mouth & 0.35 \\
        Torso & 0.24 \\
        Concreteness & 0.73 \\
        \bottomrule
    \end{tabular}
  \caption{Average sensory strength score for human-written stories along each axis.}
  \label{tab:avgHumanSenses}
\end{wraptable}

Every model studied differs significantly at $\alpha \leq 0.05$ from human usage of sensory language along at least ten out of twelve sensory axes; most models differ significantly along all twelve. 
This strongly demonstrates that models do not mimic human levels of sensory language use. We note that the average differences are relatively small compared to the average sensory values along each axis for human stories (Table \ref{tab:avgHumanSenses}); however, the t-tests confirm that the differences are nonetheless significant. 

We also find clear differences in how models from different families deviate from human sensory language usage.
Models from all families except GPT and Gemini use sensory language significantly less than humans along nearly ever axis (Figure \ref{fig:sensoryAxes}). 
In contrast, Gemini models use significantly more sensory language than human writers along most axes and GPT models fall between the Gemini models and models from the other families.

\begin{figure*}[t]
    \centering
    \includegraphics[width=0.6\linewidth]{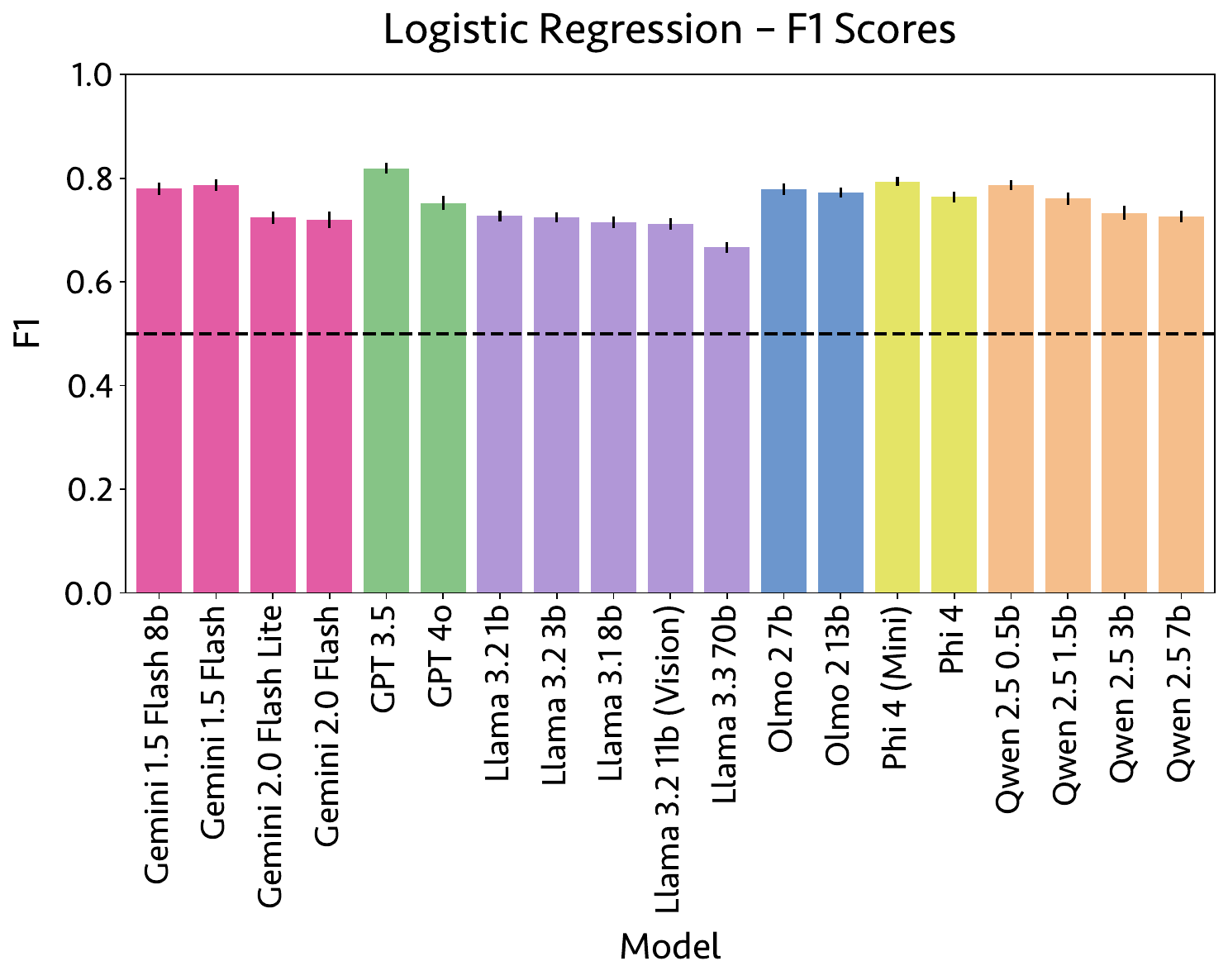}
    \caption{The average performance (F1) of 100 logistic regression models in distinguishing between texts written by humans and texts written by each model. Models are trained with different 50:50 training and test splits. The error bars represent the standard deviation over all 100 model runs and the dotted line marks expected random performance.}
    \label{fig:logRegPerformance}
\end{figure*}

There are particularly clear divides between the model families along the visual and concreteness axes; the Gemini models and GPT-4o use significantly more visual and concrete language than humans whereas the other models use significantly less. 
To exemplify these differences, we provide the first two sentences of the responses to a single prompt\footnote{``Every human that's ever lived has met God.
He takes the shape of a mailman, teacher, store clerk or another passerby to evaluate you based on one morality test. 
Out of the 107 billion humans that's ever been alive you are the only one that figured out who he really is...''} from Gemini 2.0 Flash, one of the models that uses the most concrete and visual language compared to human writers, and Phi 4, one of the models that uses the least concrete and visual language:

\begin{displayquote}
    \textbf{Gemini 2.0 Flash:} ``The chipped ceramic mug warmed my hands, the lukewarm tea doing little to soothe the tremor in my soul. Rain lashed against the windowpane, mimicking the relentless rhythm of my thoughts.''
\end{displayquote}

\begin{displayquote}
    \textbf{Phi 4:} ``Ever since I was a child, I had an uncanny knack for noticing the peculiarities in everyday life. It was a skill that often left me feeling isolated, as if I were the only one who saw the world through a different lens.''
\end{displayquote}


The difference between model and human language usage is stronger for GPT-3.5 than GPT-4o along all but three axes: the action effectors Head, Mouth, and Hand + Arm.
This suggests that changes made between the creation of these two models may have reduced the difference between the extent of human and model sensory language use.
In particular, the multi-modal training and capabilities of GPT-4o may move its sensory language use closer to human levels.

\begin{figure}[t]
    \centering
    \begin{minipage}{0.49\textwidth}
        \centering
        \includegraphics[width=\linewidth]{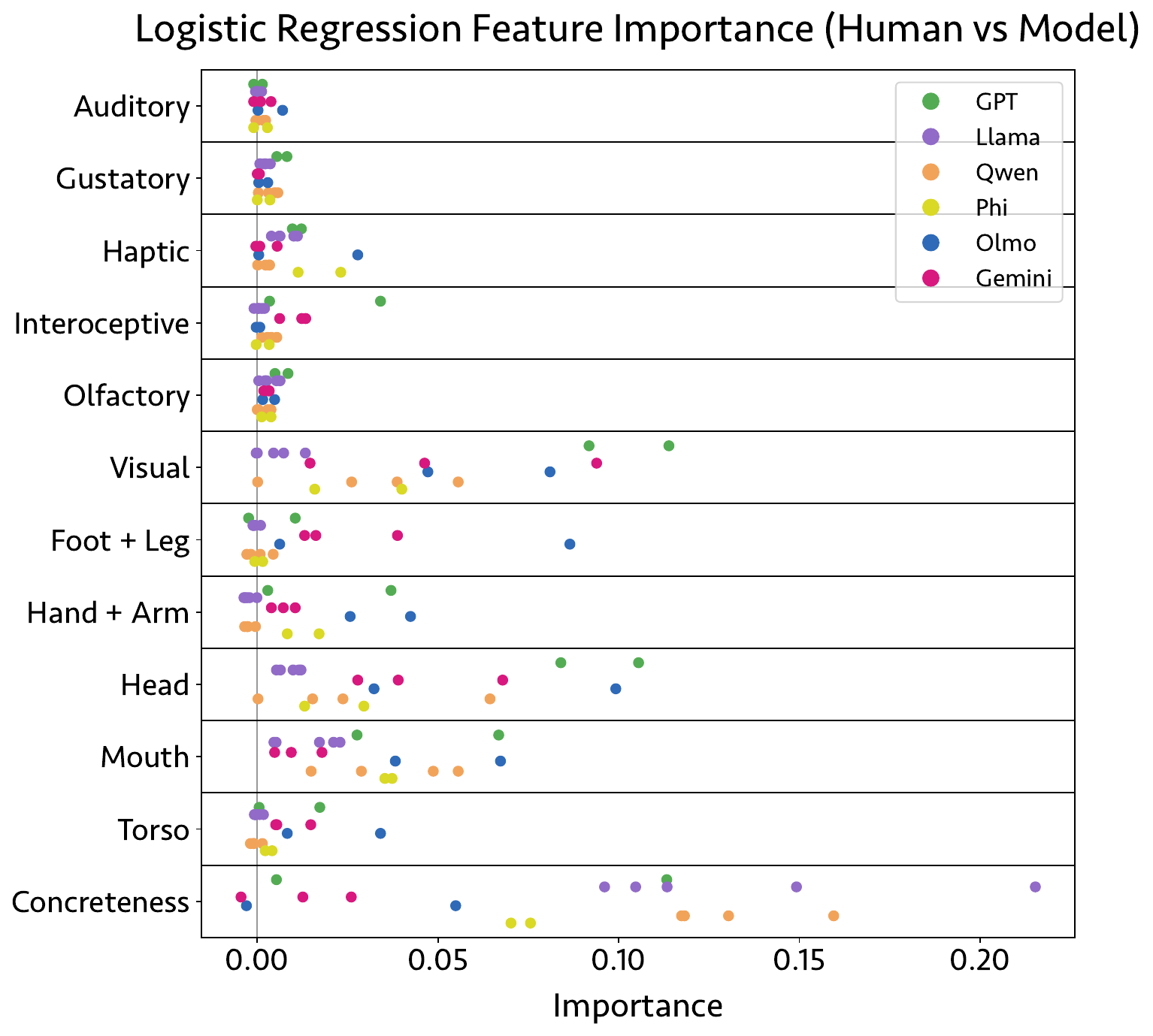}
        \caption{The average feature importance from each logistic regression model trained to distinguish between models and humans. Each dot represents an average over 100 models.}
        \label{fig:logRegImps}
    \end{minipage}
    \hfill
    \begin{minipage}{0.49\textwidth}
        \centering
        \small
        \begin{tabular}{lcc}
            \toprule
            \textbf{Sensory Axis} & \textbf{Mean Diff.} & \textbf{T-Test Stat.} \\
            \midrule
            Auditory & 0.00065 & 0.92 \\
            Gustatory & 0.00060 & 1.46 \\
            Haptic* & 0.01747 & 27.72 \\
            Interoceptive* & -0.01114 & -13.74 \\
            Olfactory* & 0.00303 & 7.13 \\
            Visual* & 0.04196 & 40.79 \\
            Foot + Leg* & 0.00930 & 19.92 \\
            Hand + Arm* & 0.02353 & 36.46 \\
            Head* & 0.01165 & 15.80 \\
            Mouth* & -0.00216 & -3.40 \\
            Torso* & 0.00877 & 19.81 \\
            Concreteness* & 0.03664 & 32.48 \\
            \bottomrule
        \end{tabular}
      \captionof{table}{Comparisons of sensory language usage between rejected and chosen responses in the Anthropic RLHF dataset \citep{bai2022training}. Negative values mean the chosen responses used more sensory language and vice versa. Axes for which there are significant differences at $\alpha=0.05$ are marked with a star (*).}
      \label{tab:anthropicSenses}
    \end{minipage}
\end{figure}

We additionally find that logistic regression models are able to distinguish between texts written by humans and each model with well above random accuracy (Figure \ref{fig:logRegPerformance}), further confirming that the sensory language use of humans and each model differs considerably. 
The logistic regression models tend to perform worse when distinguishing between human texts and those written by larger and newer models in a family, suggesting that larger models' sensory language use may be more similar to humans. By further examining the average importance of each sensory feature used by the logistic regression models, we see that the visual and concreteness axes are again frequently important (Figure \ref{fig:logRegImps}). 
Overall, action effectors appear to be more important than the perceptual modalities for the logistic regression models, although the differences between human and model usage of these axes are not necessarily more significant (Figure \ref{fig:sensoryAxes}).

\section{Probing for Sensory Language}

While our results suggest contemporary instruction-tuned language models do not replicate human patterns of sensory language usage, this does not mean LLMs fail to capture the concept during pre-training.
We can probe for this linguistic knowledge by training regression models to identify sensory language from language models' latent representations.
Researchers commonly train such simple linear models, or linear probes, to identify whether language models implicitly learn linguistic phenomena during pre-training \citep{shiDoesStringBasedNeural2016, tenneyBERTRediscoversClassical2019, marksGeometryTruthEmergent2024}.

In this experiment, we probe for each sensory axis by first calculating sequence-level sensory values for all 272,600 human-written stories from the GPT-WritingPrompts dataset. 
We pass each story through five models selected for language model type (MLM, seq2seq, and CLM), recency and ubiquity in interpretability research: BERT \citep{devlinBERTPretrainingDeep2018}, RoBERTa \citep{liuRobertaRobustlyOptimized2019}, T5 \citep{raffelExploringLimitsTransfer2023}, GPT-2 \citep{radford_language_nodate}, and Qwen 2.5 0.5b.
We collect embeddings from BERT and RoBERTa by extracting the embedding of the \texttt{CLS} token for every layer and
from T5 (encoder), GPT, and Qwen by taking the mean hidden state for every layer as a proxy for pooled embeddings.
This process yields 3,543,800 embeddings for a 12-layer transformer when including the embedding layer. 

We then tag every embedding with the corresponding sensory values observed for the sequence from which it was generated and bin the embeddings into train and test sets on a per-layer basis with a ratio of 80:20. 
Because our target sensory values are selected from the range [0,5] we use $\ell_2$-regularized ridge regression models to predict them. We allow the ridge regression implementation distributed with \texttt{scikit-learn} to automatically select a solver. For each combination of language model and sensory axis, we train a regression model to predict sensory values given a list of passages.
We then calculate the $R^2$ between these predicted values and the ground truth produced by our lexicons of interest.
Higher $R^2$ values indicate a given sensory axis is better represented in the latent representations of a given language model.
We further select the best performing $\alpha$ for each model by repeating the training process five times while incrementing $\alpha$ by 0.20, from 0 to 1. 
We finally select the best performing ridge regression model from this set, per probe.

Global probe performance is varied according to layer, model, and sensory axis.
The best performing probes achieve $R^2\approx0.85$ when predicting concreteness in deeper model layers, a result corresponding with prior work in BERTology suggesting LLMs resolve syntactic features early in processing \citep{tenneyBERTRediscoversClassical2019}.
The probes for most sensory axes achieve $0.3 \leq R^2 \leq 0.6$ across most layers of all models, including auditory, gustatory, and haptic.
The least predictable sensory axis is the torso action effector, with no probe achieving a $R^2 > 0.17$. 
Nonetheless, these results demonstrate that even smaller, older LLMs are capable of representing most sensory axes, which suggests that the difference between LLM and human use of sensory language is not due to their inability to recognize its usage.

\begin{wraptable}{r}{0.65\textwidth}
    \small
  \centering
  \begin{tabular}{lcc}
    \toprule
        \textbf{Model} & \textbf{Probe/Model} & \textbf{Anthropic/Model}\\
        \midrule
        Gemini 1.5 Flash 8b & -0.54 & -0.46 \\
        Gemini 1.5 Flash & -0.70* & -0.69* \\
        Gemini 2.0 Flash & -0.76* & -0.75* \\
        Gemini 2.0 Flash-Lite & -0.74* & -0.75* \\
        \midrule
        GPT-3.5 & 0.15 & 0.32 \\
        GPT-4o & -0.52 & -0.52 \\
        \midrule
        Llama 3.2 1b & 0.74* & 0.86* \\
        Llama 3.2 3b & 0.73* & 0.82* \\
        Llama 3.1 8b & 0.71* & 0.79* \\
        Llama 3.2 11b (Vision) & 0.71* & 0.81* \\
        Llama 3.3 70b & 0.69* & 0.92* \\
        \midrule
        OLMo 2 7b & 0.46 & 0.64* \\
        OLMo 2 13b & 0.27 & 0.51 \\
        \midrule
        Phi 4 (mini) & 0.69* & 0.76* \\
        Phi 4 & 0.78* & 0.83* \\
        \midrule
        Qwen 2.5 0.5b & 0.76* & 0.84* \\
        Qwen 2.5 1.5b & 0.59* & 0.57 \\
        Qwen 2.5 3b & 0.64* & 0.61* \\
        Qwen 2.5 7b & 0.66* & 0.65* \\
        \bottomrule
    \end{tabular}
  \caption{Correlations between average differences along each sensory axis for each model (human minus model) and (column 1) average linear probe performance or (column 2) differences in sensory language for the Anthropic RLHF dataset (rejected minus chosen). Results which are significant at $\alpha=0.05$ are marked with a star (*).}
  \label{tab:anthropicModelCorrs}
\end{wraptable}

Differences between the language models indicate newer and longer-trained models yield better performing linear probes, suggesting extended training time results in sensory values being more coherently embedded in latent representations.
The degree to which our probes resolve each sensory axis is consistent across all models with an inter-model Pearson's $R \approx 0.98$ for any two given models, indicating that all models resolve the same sensory axes to the same degree, despite variations in absolute performance.
We then quantify the relative accuracy with which the models are able to represent each sensory axis using the average probe performance across all models and all layers.

We observe significant correlation between the average probe performance per sensory axis and the human-model differences reported in Section \ref{sec:comparing sensory language}. 
We find $R\geq0.45$ for all models except for the Gemini models, GPT 3.5, and Olmo 2 13b (Table \ref{tab:anthropicModelCorrs}, Probe/Model column).
This means that the more a sensory axis is represented in latent representations of the five probed models, the \textit{less} most models use that language.
In other words, the facets that are most easily recognized by models deviate the most from human usage.

\section{Impact of RLHF Training}

We next consider whether downstream instruction tuning may be responsible for the differences between human and model sensory language usage. To explore the effect of reinforcement learning from human feedback (RLHF) on model behavior, we draw on the popular Anthropic RLHF dataset \citep{bai2022training}. 
This dataset contains 44,848 paired continuations of human-model interactions, of which one is considered a favorable response (labeled `chosen') and the other is considered unfavorable (labeled `rejected'). 
Although we do not know which datasets were used in training and finetuning many of the propriety models used in this study, we anticipate many will have trained on the Anthropic dataset or something similar. 
At a minimum, Microsoft has disclosed that it used the Anthropic dataset in training the Phi 4 models \citep{abdin2024phi}.

In this experiment, we compare how sensory language is used in the rejected and chosen responses from the Anthropic RLHF dataset. 
Because the responses are paired, we are able to treat them much like we do each set of human and model prompt responses above. 
We first generate the normalized IDF scores for all of the lemmas in this dataset. 
We then extract the parts of each paired exchange that differ between the rejected and chosen responses (the assistant's last reply).
We finally measure the difference in sensory language use between the rejected and chosen assistant responses, again producing the average difference in sensory strength along each axis and the t-test statistics comparing sensory content in each pair of responses.
This process reveals there is a significant difference in the strength of sensory language use between the rejected and chosen model responses along all axes except for auditory and gustatory (Table \ref{tab:anthropicSenses}). 
Of the significant differences, the rejected responses use more of each kind of sensory language except for interoception and mouth action effectors. 

To probe whether a relationship may exist between training with this dataset and models' non-human use of sensory language, we examine the correlation between the average differences in human and model sensory language usage (human minus model) and the average differences in rejected versus chosen responses' sensory language usage (rejected minus chosen). 
We find that significant correlations between these values exist for most models, but that again the correlations differ considerably between models from different families (Table \ref{tab:anthropicModelCorrs}, Anthropic/Model column). 

Average differences between the Gemini models and humans are all correlated negatively with trends in the Anthropic dataset.
Thus, along axes where the rejected Anthropic responses used more sensory language, the Gemini models tend to use more sensory language than humans. 
In contrast, the correlations between the GPT models' sensory language use and that in the Anthropic dataset are not significant.
For most models from the remaining four families --- Llama, OLMo, Phi, and Qwen --- there is a significant positive relationship between the average differences in human and model sensory language use and sensory language use in the rejected and chosen responses.
Thus, the more particular forms of sensory language were used in discouraged responses from the Anthropic dataset, the less models from these families used that language in comparison with humans. 
These correlations therefore provide evidence that RLHF training with the Anthropic dataset changes how models use sensory language, in particular discouraging the use of particular forms of sensory language.

\section{Conclusion}
Despite the constant tendency to anthropomorphize language models, they are not human and do not in any way experience embodied human senses.
But there is no \textit{a priori} reason to believe that they cannot emulate the language of embodied humans.

In this work we find that LLMs do \textit{not} replicate human patterns of embodied and sensory language use along twelve axes.
Differences between model and human behavior vary considerably by model family; whereas Gemini models use far more sensory language than humans along most axes, models from other studied families tend to use significantly less.
Probing for sensory language in model latent representations suggests LLMs are able to identify sensory language despite not using it.
Moreover, our results suggest the better a model understands a sensory axis the less likely it is to use it.
We investigated why this phenomenon may occur by examining a common RLHF dataset, which in turn revealed post-training RLHF may be discouraging models from using many forms of sensory language.

These findings suggest that learning from large corpora of human-written texts allows models to partially identify sensory language, but downstream instruction tuning frequently discourages its use.
This has implications for the use of LLMs in tasks requiring empathy and world awareness such as creative writing, robotics, and therapy.
Our results suggest instruction tuning may have unintended consequences on model behavior in non-obvious ways.
Possible side-effects should be carefully considered when designing and training models.
We urge LLM researchers to consider drawing on tasks from psycholinguistics and psychology to assess how language models are impacted by imperfect RLHF techniques.

\section*{Ethics Statement}
All data used in this study was either pre-existing or was generated from pre-existing language models. No human involvement was solicited, and no data is sensitive. LLM-generated stories are of no commercial value and do not compete with any human artists.

\section*{Reproducibility Statement}
Our work makes use of publicly available datasets. 
All models were accessed via HuggingFace.
We note all generations were conducted with a temperature $\geq0$, meaning results were subject to slight stochasticity.

\section*{Acknowledgments}
We would like to thank Axel Bax, Federica Bologna, Kiara Liu, Andrew Piper, Rosamond Thalken, Andrea Wang, Matthew Wilkens, and Shengqi Zhu for their thoughtful feedback. This work was supported in part by NEH grant HAA-290374-23, AI for Humanists.

\bibliography{colm2025_conference}
\bibliographystyle{colm2025_conference}

\end{document}